\documentclass[11pt,twocolumn]{article} 
\usepackage[pass,paperwidth=8.5in,paperheight=11in]{geometry}
\usepackage{graphicx}
\usepackage{amssymb}
\usepackage{hyperref}

\title{A Constraint Programming-based Job Dispatcher \\for Modern HPC Systems and Applications}

\author{
    Cristian Galleguillos,\textsuperscript{\rm 1,\rm 2} 
    Zeynep Kiziltan,\textsuperscript{\rm 2} 
    Ricardo Soto\textsuperscript{\rm 1}\\
    \textsuperscript{\rm 1} Pontificia Universidad Cat\'olica de Valpara\'iso, Valpara\'iso, Chile\\
    \textsuperscript{\rm 2} University of Bologna, Bologna, Italy\\
    cristian.galleguillos.m@mail.pucv.cl, zeynep.kiziltan@unibo.it, ricardo.soto@pucv.cl 
}

\newcommand{\CPnt}{PCP'19}
\newcommand{\CPHnt}{HCP'19}
\newcommand{\CPtw}{PCP'20}

\graphicspath{{figures/}}

\begin{document}

\maketitle

\begin{abstract}
Constraint Programming (CP) is a well-established area in AI as a programming paradigm for modelling and solving discrete optimization problems, and it has been been successfully applied to tackle the on-line job dispatching problem in HPC systems including those running modern applications. The limitations of the available CP-based job dispatchers may hinder their practical use in today's systems that are becoming larger in size and more demanding in resource allocation. In an attempt to bring basic AI research closer to a deployed application, we present a new CP-based on-line job dispatcher for modern HPC systems and applications. Unlike its predecessors, our new dispatcher tackles the entire problem in CP and its model size is independent of the system size. Experimental results based on a simulation study show that with our approach  dispatching performance increases significantly in a large system and in a system where allocation is nontrivial.

\end{abstract}

\section{Introduction}

\paragraph{Motivations}

\begin{figure}[!t]
    \centering
    \includegraphics[width=\columnwidth]{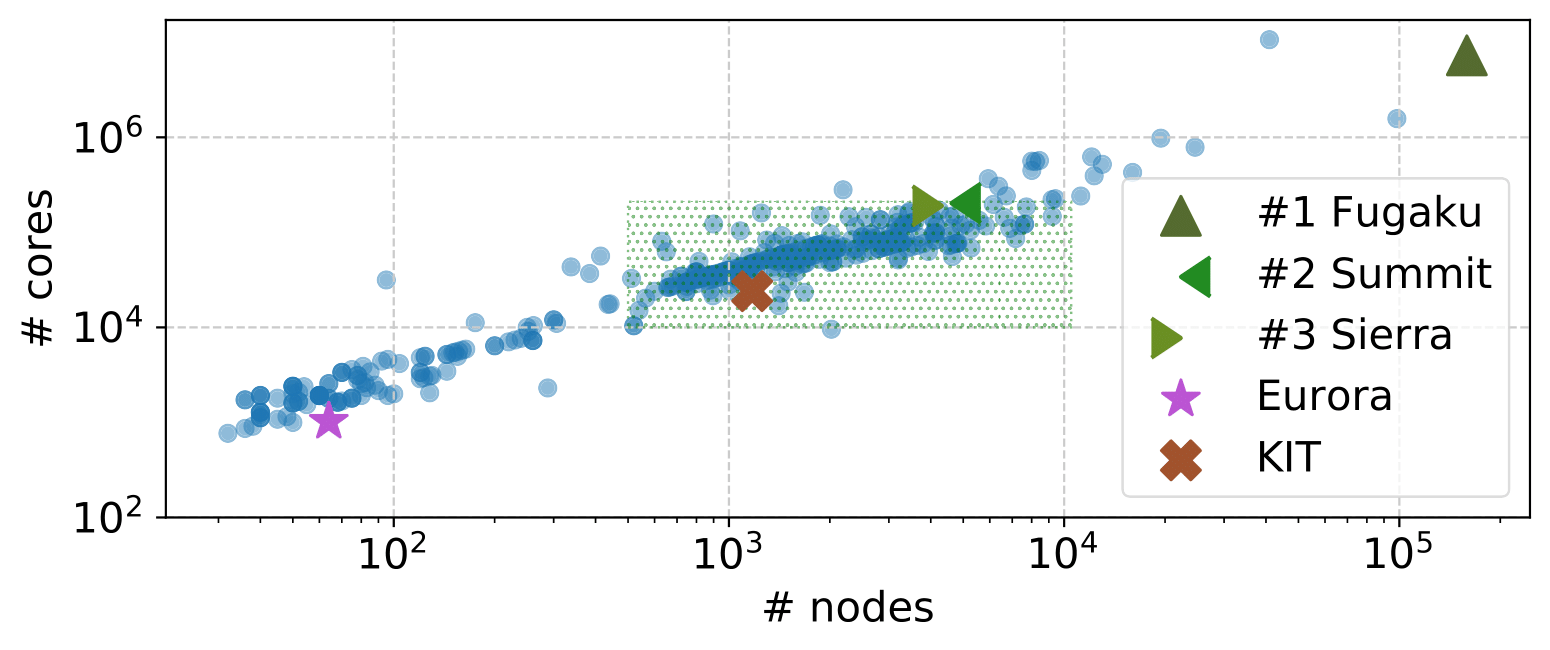}
    \caption{Size of the Eurora, KIT ForHLR II and the top 500 HPC systems.}
    \label{fig:hpc_sizes}
\end{figure}

High Performance Computing (HPC) is the application of supercomputers to solve complex computational problems  
in science, business and engineering. As such, HPC systems have become indispensable for scientific progress, industrial competitiveness, economic growth and quality of life in our modern society~\cite{ITIF,PRACE}. An HPC system is a network of computing nodes, each containing one or more CPUs and its own memory.  The next generation of HPC systems aim at reaching the exaFLOP level ($10^{18}$ floating-point operations per second). Indeed, in single or further reduced precision, which are often used in machine learning and AI applications, the peak performance of today's most powerful system Fugaku is over 1 exaFLOPS.~\footnote{\url{https://www.top500.org}} In their march towards exascale performance, HPC systems are getting larger in their number of nodes and becoming more heterogeneous in their computing resources in an effort to keep the power consumption at bay. Figure \ref{fig:hpc_sizes} shows in blue dots and green triangles the size of today's top 500 systems\footnotemark[1], where the majority has thousands of nodes. Around $30\%$ of these systems employ specialized energy-efficient 
accelerators 
such as GPUs and MICs.

Central to the efficiency and the Quality-of-Service (QoS) of an HPC system  is the  \textit{job dispatcher} which has the key role of deciding which jobs to run next among those waiting in the queue (\textit{scheduling}) and on which resources to run them (\textit{allocation}). This is an on-line decision making  in the sense that the process is repeated periodically as new jobs arrive to the system while some previously dispatched jobs are still running. Traditionally, HPC job dispatchers have been designed for compute-intensive jobs requiring days to complete. There is an increasing trend where HPC systems are being used for modern applications that employ
many short jobs ($<1$ h), such as data analytics as data is being streamed from a monitored system~\cite{reuther2018scalable}.  In such application scenarios response times are critical for acceptable user experience, hence job dispatchers need to rapidly process a large number of short jobs in making on-line decisions. 
While 
optimal dispatching 
is a critical requirement in HPC systems, the on-line job dispatching is an NP-hard optimization problem~\cite{DBLP:journals/dam/BlazewiczLK83}.

Constraint Programming (CP) is a paradigm for modelling and solving discrete optimization problems \cite{CP-handbook}, with successful applications in science, business and industry. For instance, the scientific exploratory experiments of  Philae, the first robot-lab that landed on the surface of a comet in 2014, were scheduled using CP~\cite{Simoninetal2015}. 
Though CP has its roots in logic programming and the constraint satisfaction area of AI, the last decades have witnessed its fruitful cross-fertilization with related disciplines, such as operations research, and with other areas of AI such as SAT, heuristic search, and more recently machine learning. The on-line job dispatching problem in HPC systems can naturally 
be expressed as a job scheduling and resource allocation problem for which CP has a long track of success~\cite{BAPTISTE2006761}. 



\paragraph{Related work}
\cite{DBLP:conf/cp/GalleguillosKSB19} presented two CP-based on-line job dispatchers for HPC systems, which we here refer to as \CPnt\ and \CPHnt. These dispatchers are built on previous CP-based dispatchers \cite{DBLP:conf/cp/BartoliniBBLM14,DBLP:conf/cp/BorghesiCLMB15} and are redesigned for satisfying the challenges of systems running modern applications that employ many short jobs and that have strict timing requirements.
A simulation study \cite{DBLP:conf/cp/GalleguillosKSB19} based on a workload trace collected from the 
Eurora system \cite{DBLP:conf/ics/Cavazzoni12} reveals that \CPnt\ and \CPHnt\ yield substantial improvements over the original dispatchers \cite{DBLP:conf/cp/BartoliniBBLM14,DBLP:conf/cp/BorghesiCLMB15} and provide a better QoS compared to Eurora's dispatcher \cite{DBLP:conf/jsspp/Henderson95}, which is a part of the commercial workload management system PBS Professional \cite{pbs}. 

\CPnt\ and \CPHnt\ play significant roles in the adoption of an AI-driven technology in the workload management of HPC systems, yet they have limitations which may hinder their practical use in today's systems. \CPnt\ is not scalable to large systems composed of thousands of nodes. This is because in \CPnt\ the entire problem is tackled 
using CP, and the number of decision variables increases proportionally to the number of nodes and the possible allocations of jobs in each node. Figure \ref{fig:hpc_sizes} shows where a system like Eurora stands compared to today's top 500 systems.  As we will show in our experimental results,  \CPnt\ cannot be used in a larger system like KIT ForHLR II\footnote{\label{kit}\url{https://www.scc.kit.edu/dienste/forhlr2.php}}, whose size is comparable to that of the majority of the top systems.  

In \CPHnt, the problem is decomposed and solved in two stages.  First, the scheduling problem is addressed 
using CP by treating the resources of the same type across all nodes as a pool of resources.
Then the allocation problem is solved with a heuristic algorithm using the best-fit strategy~\cite{DBLP:books/daglib/0036999}, while fixing any inconsistencies introduced in the first phase. Without an allocation model on the CP side, the number of decision variables drops dramatically and \CPHnt\ can scale to larger systems like KIT ForHLR II. However, the decoupled approach may result in several iterations between the two stages when allocation is nontrivial, for instance when many jobs demand the scarce resource types in an heterogeneous system \cite{DBLP:conf/supercomputer/NettiGKSB18}, 
or when power-aware allocation is required to limit power consumption \cite{DBLP:conf/cp/BorghesiCLMB15}. 
This in turn could decrease dispatching performance, as we will show in our experimental results with the  heterogeneous system Eurora.  The advantage of tackling the entire problem in CP is that scheduling and allocations decisions are made jointly and complex allocation constraints 
 that emerge from the needs of today's systems can be integrated in the CP model. 

\paragraph{Contributions}

We exploit the strengths of \CPnt\ and \CPHnt\ to overcome their limitations. We present a new CP allocation model where the number of variables is system size independent. We combine this model with the CP scheduling model common to  \CPnt\ and \CPHnt\ and devise a tailored search algorithm. Our contributions are (i) a novel CP-based online job dispatcher (\CPtw) suitable for modern HPC systems and applications and (ii) experimental evidence of the benefits of \CPtw\ over \CPnt\ and \CPHnt\ supported by a simulation study based on workload traces collected from the Eurora and KIT ForHLR II systems. 



\paragraph{Organization}
The rest of the paper is organized as follows. In Section~\ref{sec:bck}, we introduce the on-line job dispatching problem in HPC systems, and describe briefly the CP scheduling and allocation models of \CPnt\ as we will later use the same scheduling model in \CPtw\ and contrast the allocation model with ours. In Section~\ref{sec:new_pcp}, we present our new CP allocation model and search algorithm. 
In Sections~\ref{sec:exp-study} and \ref{sec:expes}, we detail our simulation study and present our results. We conclude and describe the future work in Section~\ref{sec:conclusions}. 

\section{Formal Background}
\label{sec:bck}
\subsection{On-line job dispatching problem in HPC systems}
\label{sec:problem}

A \textit{job} is a user request in an HPC system and consists of the execution of a computational application over the system resources.  A set of jobs is a \textit{workload}.  A job has a name, required resource types (cores, memory, etc) to run the corresponding application, and an \textit{expected duration} which is the maximum time it is allowed to execute on the system.  An HPC system typically receives multiple jobs simultaneously from different users and places them in a \textit{queue} together with the other waiting jobs (if there are any). The \textit{waiting time} of a job is the time interval during which the job remains in the queue until its execution time. 

An HPC system has $N$ nodes, with each node $n \in N$ having a capacity $cap_{n,r}$ for each of its resource type $r \in R$. Each job $i$ in the queue $Q$ has an arrival time $q_i$, maximum number of requested nodes $rn_i$ and a demand $req_{i,r}$ giving the amount of resources required from $r$ during its expected duration $d_i$. The resource request of $i$ is distributed among $rn_i$ identical job units, each requiring $req_{i,r}/rn_i$ amount of resources from $r$. A specific resource can be used by one job unit only. 
We have $rn_i=1$ for serial jobs and $rn_i>1$ for parallel jobs.  The units of a job can be allocated on the same or different nodes, depending on the system availability.
On-line job dispatching takes place at a specific time $t$ for (a subset of) the queued jobs $Q$.  The \textit{on-line job dispatching problem} at a time $t$ consists in \textit{scheduling} each job $i$ by assigning it a start time $s_i \geq t$, and \textit{allocating} $i$  to the requested resources during its expected duration $d_i$, such that the capacity constraints are satisfied: at any time in the schedule, the capacity $cap_{n,r}$ of a resource $r$ is not exceeded by the total demand $req_{i,r}$  of the jobs $i$ allocated on it, taking into account the presence of jobs already in execution.  The objective is to dispatch in the best possible way according a measure of QoS, such as with minimum waiting times $s_i - q_i$ 
for the jobs,  which is directly perceived by the HPC users. A solution to the problem is a \textit{dispatching decision}.  Once the problem is solved, only the jobs with $s_i = t$ are dispatched. The remaining jobs with $s_i > t $ are queued again with their original $q_i$. During execution, a job exceeding its expected duration is killed. It is the workload management system software that decides the dispatching time $t$ and the subsequent dispatching times.

\subsection{\CPnt\ dispatcher}\label{sec:pcp3}


\paragraph{Scheduling model}
The scheduling problem is modeled using Conditional Interval Variables (CIVs) \cite{DBLP:conf/flairs/LaborieR08}. A CIV $\tau_i \in \tau$ represents a job $i$ and defines the time interval during which $i$ runs. At a dispatching time $t$, there may already be jobs in execution which were previously scheduled and allocated. We refer to such jobs as running jobs. The scheduling model considers in the $\tau$ variables both the running jobs and a subset $\bar{Q} \subseteq Q$ of the queued jobs that can start execution as of time $t$.   
The properties $s(\tau_i)$ and $d(\tau_i)$
correspond respectively to the start time and the duration of the job $i$. 
Since the actual runtime (real) duration $d^r_i$ of a running or queued job $i$ is unknown at the modeling time, \CPnt\ uses an \emph{expected duration} $d_i$ for $d(\tau_i)$, which is supplied by a job duration \emph{prediction method}. For the queued jobs, we have $d(\tau_i)=d_i$. For the running jobs instead, $d(\tau_i)=max(1, s(\tau_i)+d_i-t)$ taking into account the possibility that $d_i < d^r_i$ due to underestimation. 
While the start time of the running jobs have already been decided, the queued jobs have $s(\tau_i) \in \left[t, eoh\right]$, where $eoh$ is the end of the worst-case makespan calculated as $t + \sum_{\tau_i} {d(\tau_i)}$.

The capacity constraints are enforced via the \texttt{cumulative} constraint~\cite{DBLP:conf/jfplc/AggounB92} as $\texttt{cumulative}([s(\tau_i)], [d(\tau_i)], [req_{i,r}], 
Tcap_{r} )$,
for all $n \in N$ and for all $r \in R$, with $Tcap_{r}=\sum_{n}^{N} cap_{n,r}$.
The objective function minimizes the total job slowdown $\sum_{\tau_i}{\frac{s(\tau_i)-q_i+d(\tau_i)}{d(\tau_i)}}$. The search for solutions focuses on the jobs with highest priority where the \emph{priority} of a job $i$  is its slowdown $\frac{t-q_i+d(\tau_i)}{d(\tau_i)}$ at the dispatching time $t$.   

\paragraph{Allocation model}

The allocation model replicates each $\tau_i$ variable $p_{i,n}$ times for each $n \in N$, where $p_{i,n}= min(rn_i, \min_{r \in R}{\lfloor {\frac{cap_{n,r}}{req_{i,r}/rn_i} \rfloor}})$ giving the minimum times a job unit can fit on $n$. Such a variable $u_{i,n,j}$  represents a possible allocation of a job unit $j$ of $i$ on node $n$ and has $s(u_{i,n,j})=s(\tau_i)$ and $d(u_{i,n,j})=d(\tau_i)$. To define the allocation, the model relies on the execution state property ($x$) of CIVs. We have  $x (u_{i,n,j}) \in \left[0, 1\right]$, meaning that it can be present or not in the solution. Instead for the scheduling variables we have $x(\tau_i) = 1$ because all of them need to be scheduled and thus be present in the solution. 
The model uses the \texttt{alternative} constraint~\cite{DBLP:conf/flairs/LaborieR08} to restrict the number of variables in $\cup_{n \in N}[x(u_{i,n,j})]$ present in the solution to be the maximum number of requested nodes $rn_i$, that is $\sum_{n \in N}^{}\sum_j{x(u_{i,n,j}) = rn_i}$ with $s(\tau_i)=s(u_{i,n,j})$ iff $x(u_{i,n,j}) = 1$. Additionally, the capacity constraints are enforced for each $n \in N$ and for each $r \in R$ as $\texttt{cumulative}([s(u_{i,n,j})], [d(u_{i,n,j})], [req_{i,r}/rn], cap_{n,r} )$.

A drawback of this model is its number of variables. While the scheduling model has $|\bar{Q}|$ variables, the allocation model has $\sum_{i \in \bar{Q}} \sum_{n \in N} p_{i,n}$ variables, which increases proportionally to $N$ (i.e., system size). 
Minimum $1 + |N|$ variables are needed to model a serial job. Parallel jobs require even more variables which may create difficulty in big systems with many parallel jobs.  

\section{\CPtw: a New CP-based Job Dispatcher}\label{sec:new_pcp}

Our new dispatcher \CPtw\ imports the scheduling model, the objective function and the job priorities of  \CPnt\ and contains a new allocation model with  $|\bar{Q}| + \sum_{i \in \bar{Q}} rn_i * |R|$ variables, which is system size independent. The number of variables thus depends mainly on the workload, with a variable number of requested nodes for each job $i$ multiplied by the number of resource types in the system which has a small value. In the following, we first present the allocation model and then describe how we search on the scheduling and the allocation variables. 

\paragraph{Allocation model} 

\begin{figure}[!ht]
 \centering
    \includegraphics[width=\columnwidth]{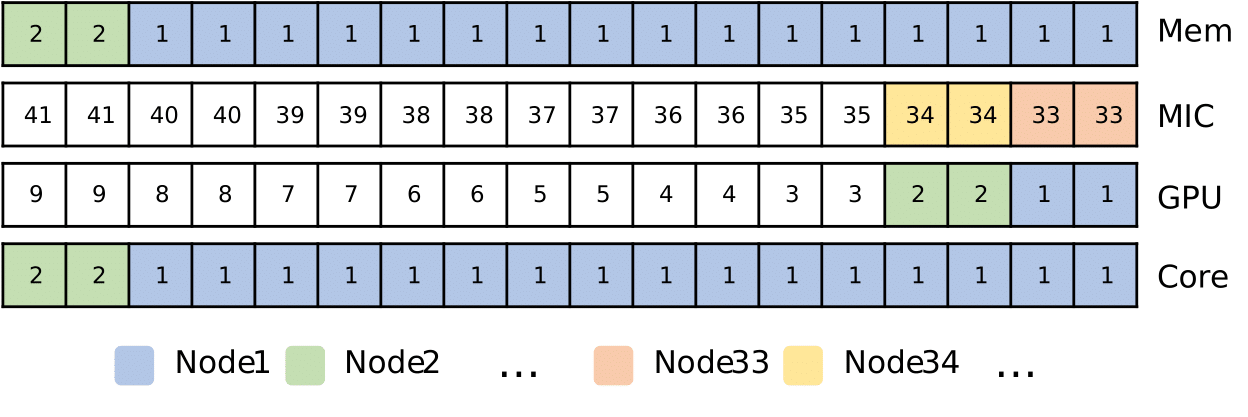}
 \caption{Node mapping on the Eurora system.}\label{fig:eurora_mapping}
\end{figure}

In this new model, we represent the system in a way to 
emphasize the resources instead of the nodes as in the previous model. 
We consider all the resources of a certain type $r$ in an ordered list by following the sequence of the nodes. This is exemplified in Figure~\ref{fig:eurora_mapping} which represents partially the Eurora system composed of 64 nodes. Each node has 16 cores and 16 GM memory, additionally the first 32 nodes have 2 GPUs, and  the next 32 has 2 MICs instead of GPUs. In the figure, the line labelled as GPU, for instance, lists all the GPU resources available in the system. There are in total $2*32$ GPUs, the first two in the list are from the first node, the third and the forth from the second node and so on. Each position in a list thus refers to a specific resource of type $r$ in a node $n$.

\begin{figure}[!t]
 \centering
     \includegraphics[width=.9\columnwidth, angle=0]{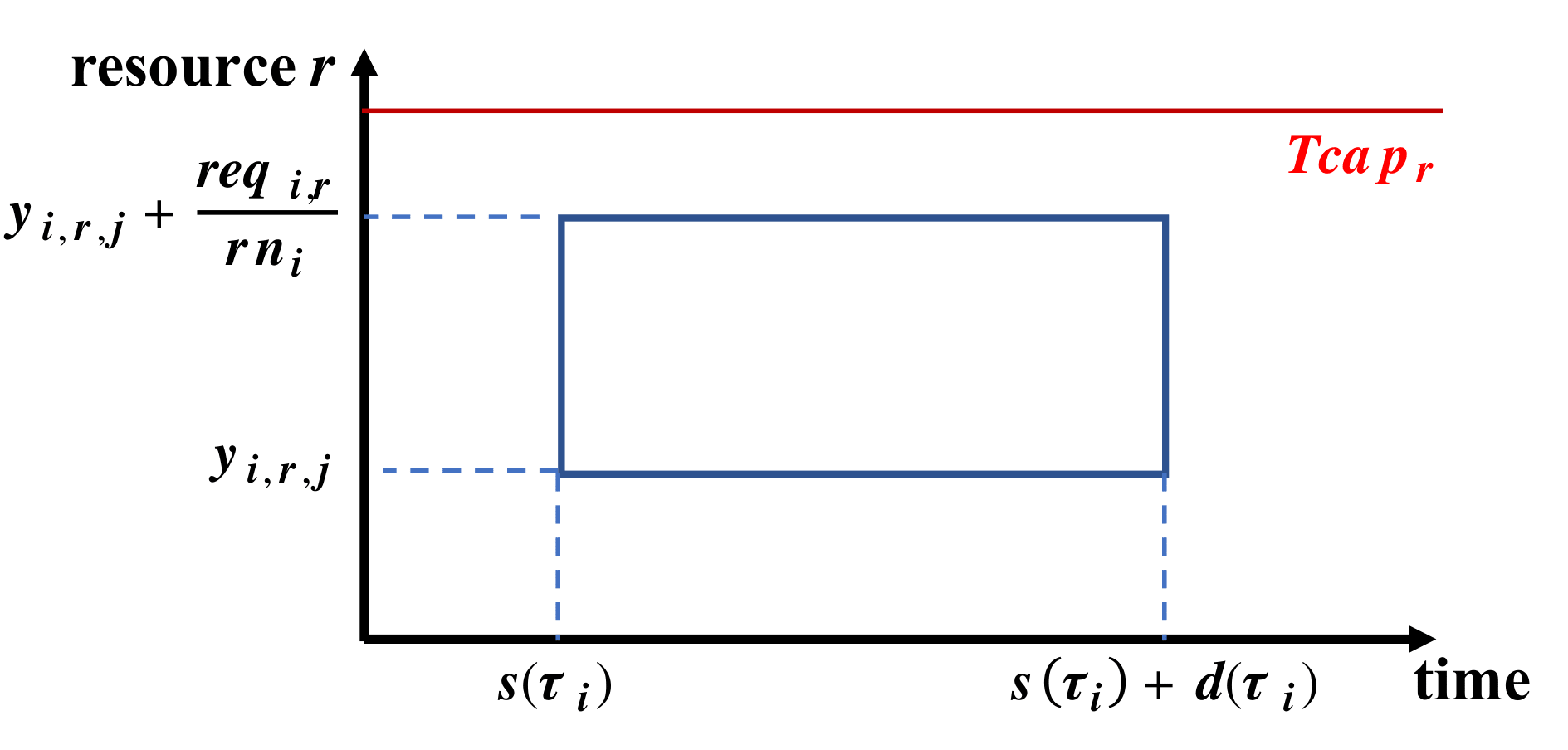}
 \caption{Representation of the allocation of a job unit on a resource type as a box.}\label{fig:job_unit_repr}
\end{figure}

\noindent
This representation allows to model the position of a job unit in a timeless way, however, it can be easily transformed to a 2-dimensional representation considering also the time, as depicted in  Figure~\ref{fig:job_unit_repr}, where the y-axis gives the available positions to allocate a job unit on a resource type $r$, and the x-axis gives the time interval during which the job unit will consume the resource. In detail, the allocation of a unit $j$ of a job $i$ on a resource type $r$ is represented as a box. The vertices of the box are defined by the variables in the origin: $s(\tau_i)$ which is  the starting time of the job $i$ and $y_{i,r,j}$ which is the starting position of the allocation of the job unit in the new system representation. The box spans from the origin to the expected duration $d(\tau_i)$ in the x-axis, instead in the y-axis to $req_{i,r}/rn_{i}$ which is the amount of resource required by the job unit. 
As for the domains of the variables, we have $D(y_{i,r,j}) = \left[1, Tcap_r\right]$, where $Tcap_r=\sum_{n \in N}cap_{n,r}$. The domain of the starting time remains the same as in the scheduling model, that is $D(s(\tau_i)) = \left[t,eoh\right]$.

To enforce that 
a resource can be used by one job unit only, we forbid the boxes to overlap via the \texttt{diffn} 
constraint. For each $r \in R$, we have  $\texttt{diffn}(\left[s(\tau_i)\right], \left[d(\tau_i)\right], \left[y_{i,r,j}\right],\left[req_{i,r}/rn_i\right])$. 
As the domain size of the $y_{i,r,j}$ variables depends on the system size and can be very large, we add implied constraints to the model  to shrink the domains. The first one regards the positions $y_{i,r,j}$ of a job $i$ on a resource type $r$ when $rn_i > 1$. We post $\texttt{alldifferent}(\left[y_{i,r,j}\right])$ to ensure that the positions are different. The other implied constraints are the classical  $\texttt{cumulative}$ constraints used together with a $\texttt{diffn}$ constraint in packing problems, as was also done in \cite{DBLP:conf/cp/SimonisO08}: 
$\texttt{cumulative}([s(\tau_i)], [d(\tau_i)], [req_{i,r}/rn_i], Tcap_r)$ and   $\texttt{cumulative}([y_{i,r,j}], [req_{i,r}/rn_i],[s(\tau_i)], eoh)$. 

Finally, we need additional constraints to guarantee that certain job units are allocated in the same node.  For that, we utilize a mapping array $map_r$ for each resource type $r$, which is based on the new representation of the system introduced earlier. The positions of $map_r$ correspond to the available resources, indexed by 1 to $Tcap_r=\sum_{n \in N}cap_{n,r}$, and each value in the array is a number corresponding to a system node.  To ensure that a unit $j$ of a job $i$ of each $r$ are allocated in the same node, we post an $\texttt{element}$ constraint, which indexes an array with a variable, as
$\texttt{element}(map_{r_1}, y_{i,r_1,j}) = \texttt{element}(map_{r_2}, y_{i,r_2,j})\ \forall r_1, r_2 \in \hat{R}$, where $\hat{R}$ is the set of the requested resource types of the unit $j$ of job $i$.  We use the $\texttt{element}$ constraint also to enforce that the covered positions spanning from $y_{i,r,j}$ to $y_{i,r,j}+req_{i,r}/rn_i$ are in the same node: $\texttt{element}(map_{r}, y_{i,r,j}) = \texttt{element}(map_{r}, y_{i,r,j} + req_{i,r}/rn_i) \ \forall r \in \hat{R} \; \textnormal{iff} \; req_{i,r}/rn_i > 0$.

\paragraph{Search}

We search on the scheduling and the allocation variables by interleaving the scheduling and the allocation assignments of a selected job. At each decision node during search, we select the job $i$ whose priority is highest and that can start first. Note that the priorities are calculated once statically at the dispatching time $t$ before search starts. We assign to $s(\tau_i)$ its earliest start time $min(D(s(\tau_i)))$. Then among the allocation variables $[y_{i,r,j}]$ of $i$, we select the one that has the minimum domain and assign it to its maximum value, by following the best-fit strategy.

\section{Experimental Study}
\label{sec:exp-study}

To evaluate the significance of our approach,  we conducted an 
experimental study by simulating on-line job submission  
two HPC systems. We dispatched the jobs using  \CPtw, \CPnt, \CPHnt, and compared them in various aspects. 

\paragraph{HPC systems and workload datasets} Our study is based on workload traces collected from two HPC systems different in size and architecture. The first is the KIT ForHLR II system\footnotemark[2], located at Karlsruhe Institue of Technology in Germany. 

The system size is comparable to the current trend (see Figure \ref{fig:hpc_sizes}) with 1,152 thin nodes, each equipped with 20 cores and 64 GB memory, along with other 21 fat nodes each containing 48 cores, 4 GPUs, and 1 TB memory. The workload dataset is available on-line\footnote{\url{https://www.cse.huji.ac.il/labs/parallel/workload/logs.html}} and contains logs for 114,355 jobs submitted during the time period June 2016--January 2018. Of all the jobs, 66.26\% are short ($<$ 1h). The second is the Eurora system ~\cite{DBLP:conf/ics/Cavazzoni12}, which was in production  at CINECA datacenter in Italy until 2015. With 64 nodes, the system size is small compared to the current trend (see Figure \ref{fig:hpc_sizes}), but the architecture is heterogeneous with each node containing 2 octa-core CPUs, 16 GB memory, and two of GPU or MIC. We use the workload dataset with which \CPnt\ and \CPHnt\ were tested in \cite{DBLP:conf/cp/GalleguillosKSB19}. 
It consists of logs over 400,000 jobs submitted during the time period March 2014--August 2015 and is dominated by short jobs, making up 93.14\% of all jobs.

\paragraph{Job duration prediction}
We derived the expected durations $d_i$ of jobs via three prediction methods. The first is a data-driven heuristic first proposed in \cite{DBLP:conf/mod/GalleguillosSKB17} and later used with \CPnt\ and \CPHnt\ during the simulation of the Eurora dataset \cite{DBLP:conf/cp/GalleguillosKSB19}. Despite being a valid alternative, this method relies on job names, a type of data omitted in the KIT and some other public datasets. We thus employed a second heuristic method that uses the run times of the last two jobs to predict the duration of the next job \cite{DBLP:journals/tpds/TsafrirEF07}.   
In both methods, the predictions are calculated on-line during the simulation and the knowledge base is updated upon job termination. The last prediction method is an oracle which gives the actual runtime (real) durations $d^r_i$ and provides a baseline during the simulation of both datasets. 

\paragraph{Simulation} 

We used the 
AccaSim workload management system simulator~\cite{DBLP:journals/cluster/GalleguillosKNS20} to simulate the 
HPC systems with their workload datasets. Each job submission is simulated by using its available data, for instance, the owner, the requested resources, and the real duration, the execution command or the name of the application executed. AccaSim uses the real duration 
to simulate the job execution during its entire duration. 
Therefore job duration prediction errors do not affect the running time of the jobs with respect to the real workload data. The dispatchers are implemented using the AccaSim directives to allow them to generate the dispatching decisions during the system simulation. 

\paragraph{Experimental setup} 

As a CP modelling and solving toolkit, we customized Google OR-Tools\footnote{\url{https://developers.google.com/optimization/}} 7.3 and ported it to Python 3.6 to implement \CPtw\ in AccaSim. As for \CPnt\ and \CPHnt, we used their publicly available implementations\footnote{\url{https://git.io/fjia1}}, and carried over their parameters to \CPtw. All experiments were performed on a CentOS machine equipped with Intel Xeon CPU E5-2640 Processor and 15GB of RAM. The source code of all the dispatchers is available at \url{https://git.io/fjia1}.

\section{Experimental Results}\label{sec:expes}

In this section, we show our experimental results. In each simulation, we compare the dispatchers' performance (in Tables \ref{table:timings-exp1} and \ref{table:timings-exp2}) in terms of (i) the average CPU time spent in generating a dispatching decision over all dispatcher invocations, including the time for modeling the dispatching problem instance and searching for a solution, and (ii) the total simulation time from the first job submission until the last job completion. We also compare the dispatchers' QoS (in Figures \ref{fig:sd_wt_big_system} and \ref{fig:sd_wt_small_system}) in terms of the average slowdown and waiting times of the jobs.  To refer to a dispatcher using a certain job duration prediction method, we append -D, -L2 or -R to the name of the dispatcher for the data-driven heuristic, the last-two heuristic and the real duration, respectively. 

\subsection{Simulation of the KIT ForHLR II workload}

\begin{figure}[!t]
    \centering
    \includegraphics[width=\columnwidth]{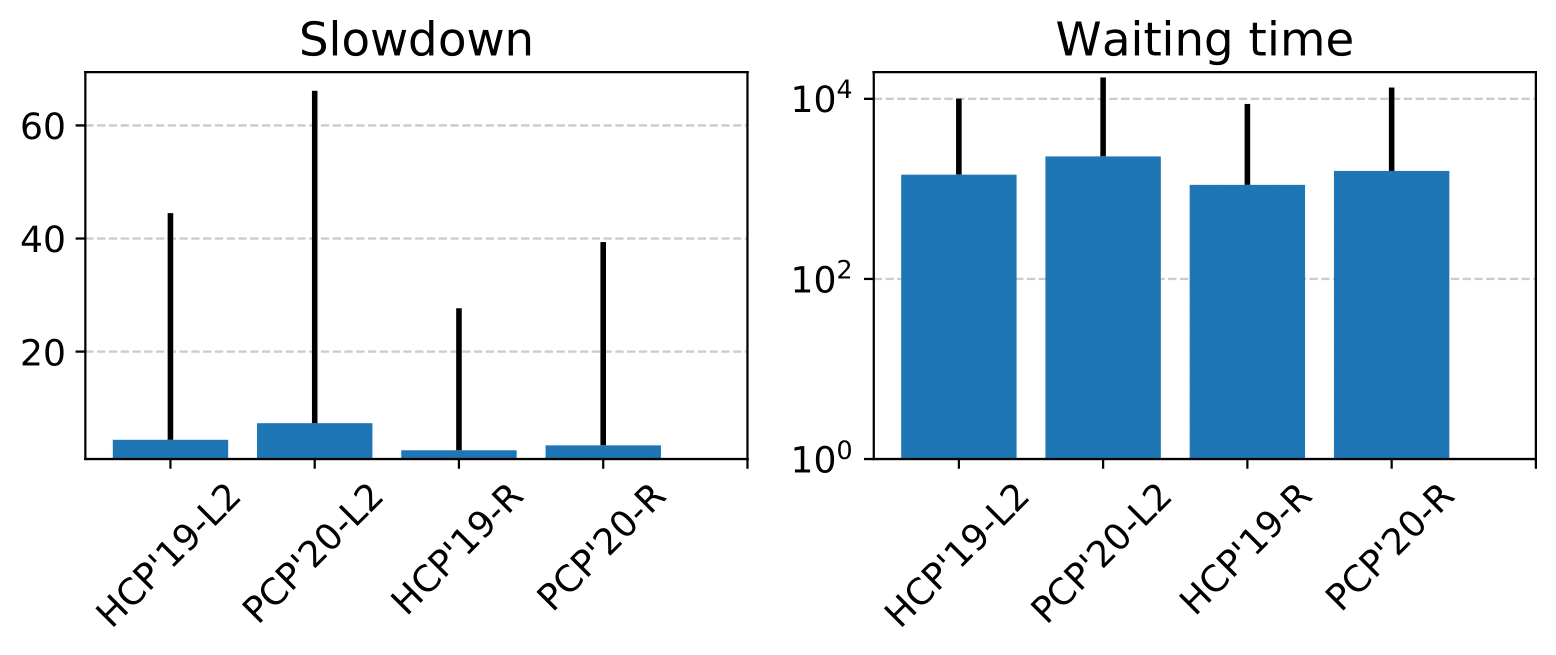}
    \caption{Average and error bars showing one std. deviation of slowdown and waiting times [s] 
    obtained from KIT.}
\label{fig:sd_wt_big_system}
\end{figure}

\CPnt\ cannot not finalize the simulation of a big system like KIT ForHLR II. At some point in time, it stops dispatching, even if new jobs are entering in the queue and the system is empty with all its resources available. This is because \CPnt\ cannot handle certain dispatching instances within the available time limit and blocks the current and the next dispatching decisions.  
\CPtw\ and  \CPHnt\ instead complete the simulation, confirming their advantage to \CPnt\ in a big system. Comparing 
\CPtw\ and  \CPHnt\ (Table \ref{table:timings-exp1} and Figure \ref{fig:sd_wt_big_system}), we see that \CPHnt\ has a much better performance than \CPtw\, and provides a slightly better QoS. This is not surprising due to the system architecture with only CPU cores and memory in 98$\%$ of its nodes. In such an homogeneous system, allocation is rather trivial. The decisions generated in the scheduling stage of \CPHnt\ are often feasible also during the allocation stage, with no need of a special allocation approach. 

\subsection{Simulation of the Eurora workload}


All dispatchers  finalize the simulation of a small system Eurora. Comparing their results (Table \ref{table:timings-exp2} and Figure \ref{fig:sd_wt_small_system}), we can clearly see the benefits of using \CPtw. In an heterogeneous system, allocation decisions are nontrivial, hence the decoupled approach of \CPHnt\ decreases significantly the dispatcher performance. We observe a further performance decrease in \CPnt\ which can be attributed to its higher number of decisions variables. While the quality of the dispatching decisions are comparable across the dispatchers (and are superior to those of the Eurora's dispatcher PBS), we note the substantial decrease in the error of average slowdown from  \CPHnt -D\ to \CPnt -D and then to \CPtw -D.

\begin{table}[!t]
\centering
\resizebox{\columnwidth}{!}{
\begin{tabular}{r|c|c}
   Dispatcher  & Avg. disp. time [ms] & Total sim. time [s]\\ \hline
   \CPHnt -L2 & 292 & 58,934 \\
   \CPnt -L2 & $\infty$ & $\infty$ \\
   \CPtw -L2 & 537 & 108,608 \\ \hline
   \CPHnt -R & 270 & 54,719 \\
   \CPnt -R & $\infty$ & $\infty$ \\
   \CPtw -R & 662 & 133,086 \\ \hline
\end{tabular}
}
\caption{Times obtained from the KIT ForHLR II system.}
\label{table:timings-exp1}
\end{table}%

\subsection{Individual instances}

An additional analysis is needed in order to quantify the reduction in the number of decisions variables obtained by going from \CPnt\ to \CPtw. During the simulation of an HPC system and its workload data, all dispatchers start with the same dispatching instance, but then they schedule and allocate jobs diversely. This in turn leads to different jobs running on different resources of the system as well as to different jobs waiting in the queue in the next dispatching time.  We cannot therefore compare the dispatchers' model size on the distinct instances they entail throughout the simulation period. To analyze the dispatchers on the same instances, we saved the instances created during the simulation of the Eurora workload  while  using \CPnt-D and \CPnt-R as a dispatcher. Each instance is created when the simulator
calls the corresponding dispatcher, and the instance is described by the  queued jobs, the running jobs and their specific allocation on the system. We obtained in total 624,564 instances.  

\begin{table}[!t]
\centering
\resizebox{\columnwidth}{!}{
\begin{tabular}{r|c|c}
Dispatcher & Avg. disp. time [ms] & Total sim. time [s]\\ \hline
   \CPHnt -D & 411  & 219,142 \\
   \CPnt -D & 565 & 301,078 \\
   \CPtw -D & 252 & 134,240 \\ \hline
   \CPHnt -R & 385 & 204,363 \\
   \CPnt -R & 512 & 272,925 \\
   \CPtw -R & 364 & 193,751 \\ \hline
\end{tabular}
}
\caption{Times obtained from the Eurora system.} 
\label{table:timings-exp2}
\end{table}%
\begin{figure}[!t]
    \centering
    \includegraphics[width=\columnwidth]{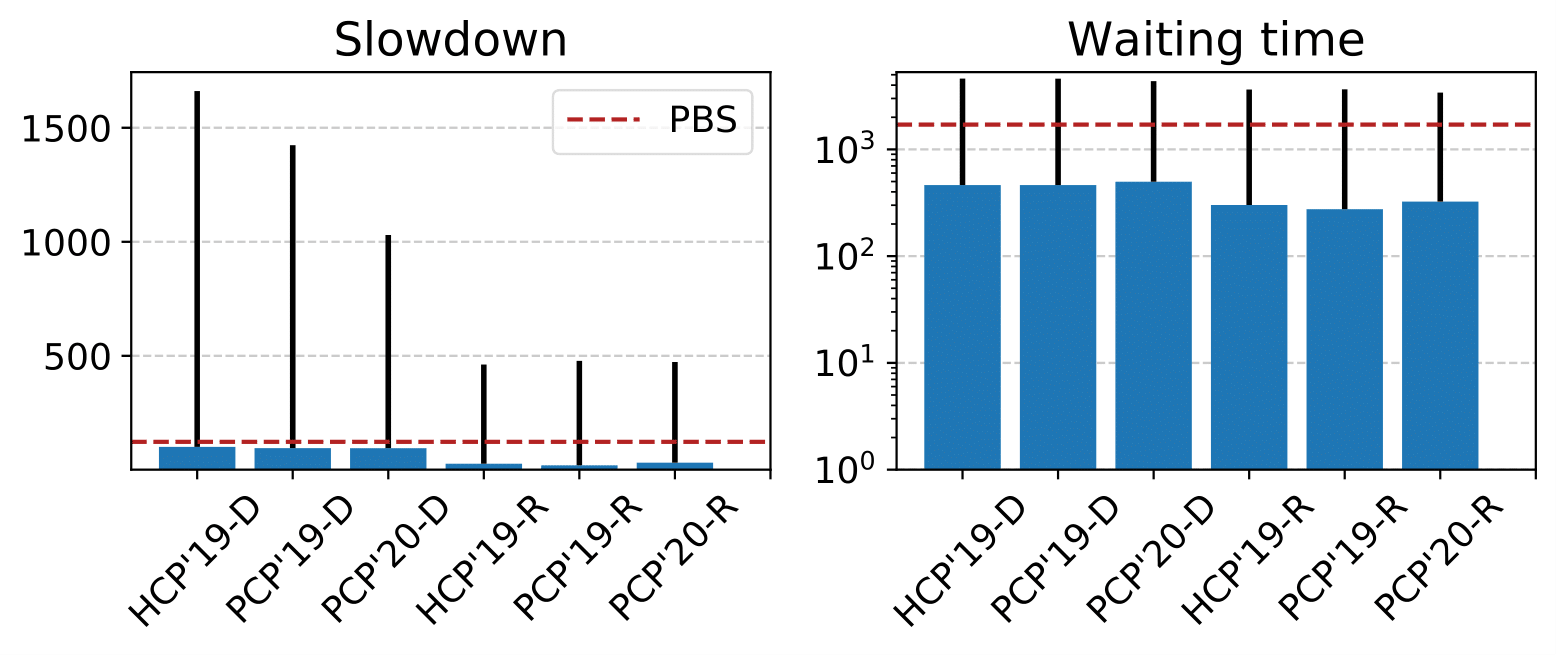}
    \caption{Average and error bars showing one std. deviation of slowdown and waiting times [s] obtained from Eurora. }
\label{fig:sd_wt_small_system}
\end{figure}

\begin{figure} [!t]
\centering
\includegraphics[width=\columnwidth]{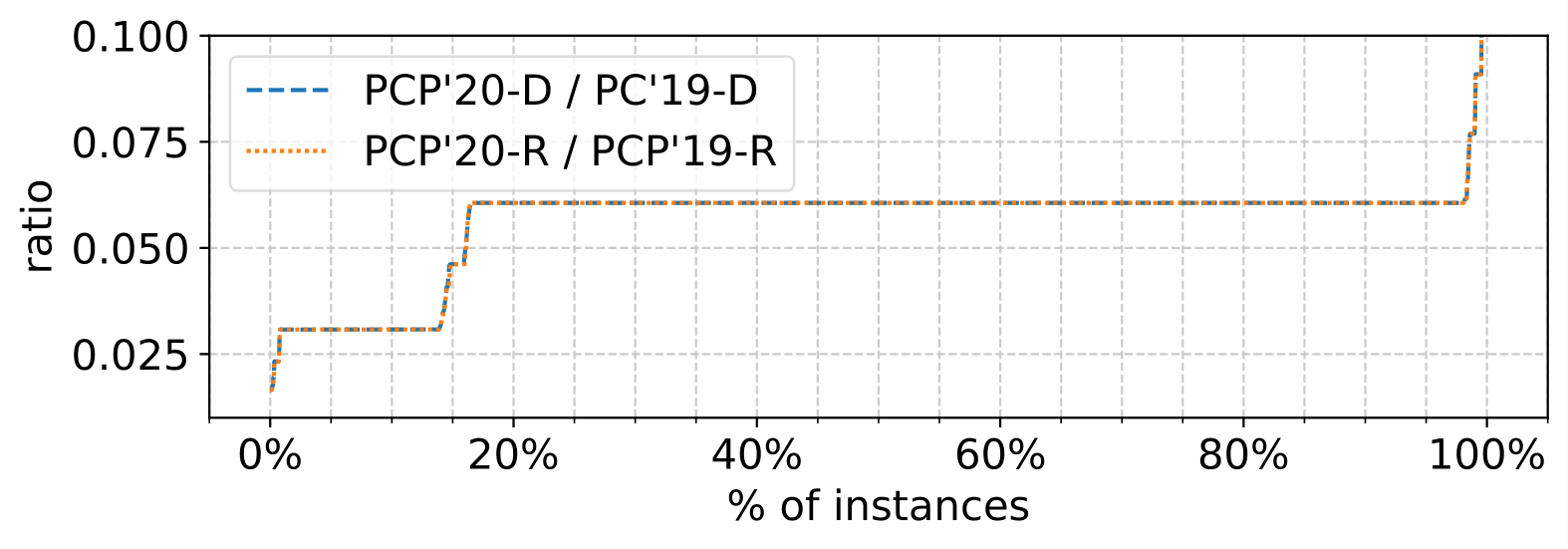}
\caption{Ratio of the number of decision variables between  \CPtw\ and \CPnt\ on the individual Eurora instances.}
\label{fig:ratio_decvars}
\end{figure}
\begin{figure} [!t]
\includegraphics[width=\columnwidth]{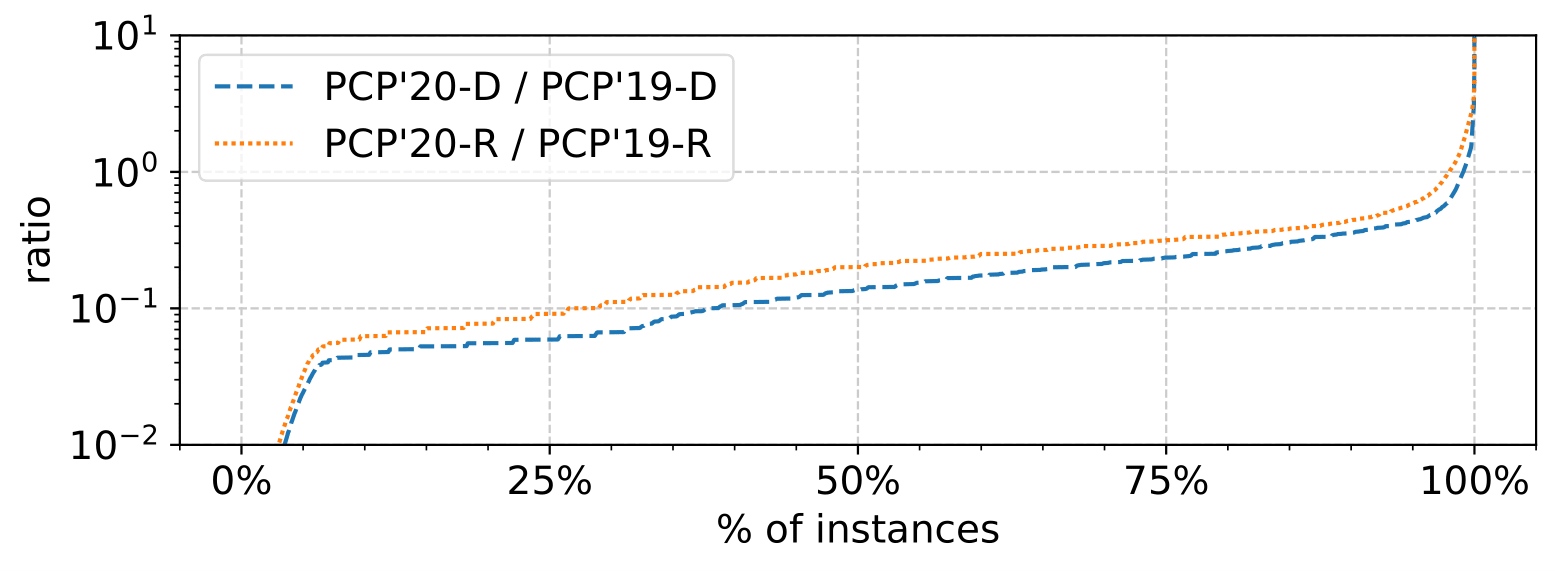}
\caption{Ratio of the dispatching time between \CPtw\ and \CPnt\  on the individual Eurora instances.}
\label{fig:ratio_time}
\end{figure}

Figure~\ref{fig:ratio_decvars} shows the ratio of the number of decision variables between \CPtw\ and \CPnt\ on each instance. 
For all instances, the ratio is below 0.1, proving the significance of the new allocation model in \CPtw. To confirm the impact on the dispatching time, we show in Figure~\ref{fig:ratio_time}  the ratio  of the dispatching time. For almost all the instances, the ratio is between 1 and 0.01, 
supporting the 
direct effect of model size on the dispatcher performance. We also analyzed the ratio of the quality of the dispatching decisions. The results (not shown for space reasons) are inline with those shown in Figure \ref{fig:sd_wt_small_system}. The ratio is 1 for the vast majority of the instances. 

\section{Conclusions and Future Work}
\label{sec:conclusions}

Constraint Programming (CP) is a well-established area in AI as a programming paradigm for modelling and solving discrete optimization problems, and it has been been successfully applied to tackle the on-line job dispatching problem in HPC systems  \cite{DBLP:conf/cp/BartoliniBBLM14,DBLP:conf/cp/BorghesiCLMB15} including those running modern applications \cite{DBLP:conf/cp/GalleguillosKSB19}. The limitations of the available CP-based job dispatchers may hinder their practical use in today's systems that are becoming larger in size and more demanding in resource allocation. In an attempt to bring basic AI research closer to a deployed application, we presented a new CP-based on-line job dispatcher for HPC systems (\CPtw). 

Unlike its predecessors, \CPtw\ tackles the entire problem in CP and its model size is independent of the system size. Experimental results based on a simulation study show that with our approach  dispatching performance increases significantly in a large system and in a system where allocation is nontrivial.

While we have used in our experiments real data representing the workload of modern applications, our conclusions are based on a simulation study which is restricted by the capabilities of the simulator. For instance, AccaSim does not add the dispatching time to the waiting times of jobs. This could be the reason why we have not observed meaningful gains in the QoS. In a real system, jobs' waiting time (and slowdown) would increase during dispatching time, therefore dispatcher performance would directly affect the QoS. We want to investigate this by modifying the simulator accordingly.  Towards our objective to deploy and evaluate a CP-based dispatcher in a real system, we plan to integrate in the model sophisticated allocation strategies, like those proposed for heterogeneous systems  \cite{DBLP:conf/supercomputer/NettiGKSB18}. Moreover, we plan to improve the search performance by breaking the symmetry introduced in the model due to the resources of the same type. 

\section*{Acknowledgements} We thank A. Bartolini, L. Benini, M. Milano, M. Lombardi and the SCAI group at Cineca for providing the Eurora data.  We also thank the School of Computer Engineering of PUCV in Chile for providing access to computing resources for simulations. C. Galleguillos has been supported by Postgraduate Grant INF-PUCV 2020.


\bibliographystyle{abbrv}
\bibliography{main}

\begin{thebibliography}{10}

\bibitem{DBLP:conf/jfplc/AggounB92}
A.~Aggoun and N.~Beldiceanu.
\newblock Extending {CHIP} in order to solve complex scheduling and placement
  problems.
\newblock In {\em JFPL'92, $1^{st}$ French Conference on Logic Programming,
  25-27 May 1992, Lille, France}, page~51, 1992.

\bibitem{pbs}
Altair.
\newblock Altair {PBS} professional (accessed september 4 2020), 2020.

\bibitem{BAPTISTE2006761}
P.~Baptiste, P.~Laborie, C.~L. Pape, and W.~Nuijten.
\newblock Chapter 22 - constraint-based scheduling and planning.
\newblock In {\em Handbook of Constraint Programming}, volume~2 of {\em
  Foundations of Artificial Intelligence}, pages 761--799. Elsevier, 2006.

\bibitem{DBLP:conf/cp/BartoliniBBLM14}
A.~Bartolini, A.~Borghesi, T.~Bridi, M.~Lombardi, and M.~Milano.
\newblock Proactive workload dispatching on the {EURORA} supercomputer.
\newblock In {\em Proceedings of Principles and Practice of Constraint
  Programming - 20th International Conference, {CP} 2014, Lyon, France,
  September 8-12, 2014.}, volume 8656 of {\em Lecture Notes in Computer
  Science}, pages 765--780. Springer, 2014.

\bibitem{DBLP:journals/dam/BlazewiczLK83}
J.~Blazewicz, J.~K. Lenstra, and A.~H. G.~R. Kan.
\newblock Scheduling subject to resource sonstraints: classification and
  complexity.
\newblock {\em Discrete Applied Mathematics}, 5(1):11--24, 1983.

\bibitem{DBLP:conf/cp/BorghesiCLMB15}
A.~Borghesi, F.~Collina, M.~Lombardi, M.~Milano, and L.~Benini.
\newblock Power capping in high performance computing systems.
\newblock In {\em Proceedings of Principles and Practice of Constraint
  Programming - 21st International Conference, {CP} 2015, Cork, Ireland, August
  31 - September 4, 2015, Proceedings}, volume 9255 of {\em Lecture Notes in
  Computer Science}, pages 524--540. Springer, 2015.

\bibitem{DBLP:conf/ics/Cavazzoni12}
C.~Cavazzoni.
\newblock {EURORA:} a european architecture toward exascale.
\newblock In {\em Proceedings of the Future {HPC} Systems - the Challenges of
  Power-Constrained Performance, FutureHPC@ICS 2012, Venezia, Italy, June 25,
  2012}, pages 1:1--1:4. {ACM}, 2012.

\bibitem{DBLP:journals/cluster/GalleguillosKNS20}
C.~Galleguillos, Z.~Kiziltan, A.~Netti, and R.~Soto.
\newblock Accasim: a customizable workload management simulator for job
  dispatching research in {HPC} systems.
\newblock {\em Cluster Computing}, 23(1):107--122, 2020.

\bibitem{DBLP:conf/cp/GalleguillosKSB19}
C.~Galleguillos, Z.~Kiziltan, A.~S{\^{\i}}rbu, and {\"{O}}.~Babaoglu.
\newblock Constraint programming-based job dispatching for modern {HPC}
  applications.
\newblock In {\em Proceeding of Principles and Practice of Constraint
  Programming - 25th International Conference, {CP} 2019, Stamford, CT, USA,
  September 30 - October 4, 2019}, volume 11802 of {\em Lecture Notes in
  Computer Science}, pages 438--455. Springer, 2019.

\bibitem{DBLP:conf/mod/GalleguillosSKB17}
C.~Galleguillos, A.~S{\^{\i}}rbu, Z.~Kiziltan, {\"{O}}.~Babaoglu, A.~Borghesi,
  and T.~Bridi.
\newblock Data-driven job dispatching in {HPC} systems.
\newblock In {\em Proceedings of Machine Learning, Optimization, and Big Data -
  Third International Conference, {MOD} 2017, Volterra, Italy, September 14-17,
  2017, Revised Selected Papers}, volume 10710 of {\em Lecture Notes in
  Computer Science}, pages 449--461. Springer, 2017.

\bibitem{DBLP:conf/jsspp/Henderson95}
R.~L. Henderson.
\newblock Job scheduling under the portable batch system.
\newblock In {\em Proceedings of Job Scheduling Strategies for Parallel
  Processing, IPPS'95 Workshop, Santa Barbara, CA, USA, April 25, 1995.},
  volume 949 of {\em Lecture Notes in Computer Science}, pages 279--294.
  Springer, 1995.

\bibitem{ITIF}
{ITIF}.
\newblock The vital importance of high-performance computing to u.s.
  competitiveness. information technology and innovation foundation. (accessed
  september 4, 2020), 2016.

\bibitem{DBLP:conf/flairs/LaborieR08}
P.~Laborie and J.~Rogerie.
\newblock Reasoning with conditional time-intervals.
\newblock In {\em Proceedings of the Twenty-First International Florida
  Artificial Intelligence Research Society Conference, May 15-17, 2008, Coconut
  Grove, Florida, {USA}}, pages 555--560. {AAAI} Press, 2008.

\bibitem{DBLP:conf/supercomputer/NettiGKSB18}
A.~Netti, C.~Galleguillos, Z.~Kiziltan, A.~S{\^{\i}}rbu, and {\"{O}}.~Babaoglu.
\newblock Heterogeneity-aware resource allocation in {HPC} systems.
\newblock In {\em Proceedings of High Performance Computing - 33rd
  International Conference, {ISC} High Performance 2018, Frankfurt, Germany,
  June 24-28, 2018}, volume 10876 of {\em Lecture Notes in Computer Science},
  pages 3--21. Springer, 2018.

\bibitem{PRACE}
{PRACE}.
\newblock The scientific case for computing in europe 2018-2026. prace
  scientific steering committee. (accessed september 4, 2020), 2018.

\bibitem{reuther2018scalable}
A.~Reuther, C.~Byun, W.~Arcand, D.~Bestor, B.~Bergeron, M.~Hubbell, M.~Jones,
  P.~Michaleas, A.~Prout, A.~Rosa, and J.~Kepner.
\newblock Scalable system scheduling for {HPC} and big data.
\newblock {\em J. Parallel Distributed Comput.}, 111:76--92, 2018.

\bibitem{CP-handbook}
F.~Rossi, P.~van Beek, and T.~Walsh, editors.
\newblock {\em Handbook of Constraint Programming}, volume~2 of {\em
  Foundations of Artificial Intelligence}.
\newblock Elsevier, 2006.

\bibitem{DBLP:books/daglib/0036999}
A.~Silberschatz, P.~B. Galvin, and G.~Gagne.
\newblock {\em Operating System Concepts, 9th Edition}.
\newblock Wiley, 2014.

\bibitem{Simoninetal2015}
G.~Simonin, C.~Artigues, E.~Hebrard, and P.~Lopez.
\newblock Scheduling scientific experiments for comet exploration.
\newblock {\em Constraints}, 20(1):77--99, 2015.

\bibitem{DBLP:conf/cp/SimonisO08}
H.~Simonis and B.~O'Sullivan.
\newblock Search strategies for rectangle packing.
\newblock In {\em Proceeding of Principles and Practice of Constraint
  Programming, 14th International Conference, {CP} 2008, Sydney, Australia,
  September 14-18, 2008.}, volume 5202 of {\em Lecture Notes in Computer
  Science}, pages 52--66. Springer, 2008.

\bibitem{DBLP:journals/tpds/TsafrirEF07}
D.~Tsafrir, Y.~Etsion, and D.~G. Feitelson.
\newblock Backfilling using system-generated predictions rather than user
  runtime estimates.
\newblock {\em {IEEE} Trans. Parallel Distrib. Syst.}, 18(6):789--803, 2007.

\end{thebibliography}

\end{document}